\documentclass[sigconf]{acmart}
\renewcommand\footnotetextcopyrightpermission[1]{}
\usepackage{algorithmic}
\usepackage{algorithm}
\usepackage{multirow}
\usepackage{tablefootnote}

\acmConference[MM '23]{Make sure to enter the correct
  conference title from your rights confirmation emai}{Oct 29 -- Nov 3, 2023}{Ottawa, Canada}

\begin{document}

\title{Sampling-Priors-Augmented Deep Unfolding Network for Robust Video Compressive Sensing}

\author{Yuhao Huang}
\affiliation{%
  \institution{Beijing Jiaotong University}
  \city{}
  \country{}}

\author{Gangrong Qu}
\affiliation{%
  \institution{Beijing Jiaotong University}
  \city{}
  \country{}}

\author{Youran Ge}
\affiliation{%
  \institution{Beijing Jiaotong University}
  \city{}
  \country{}}

\renewcommand{\shortauthors}{Huang et al.}

\begin{abstract}
    Video Compressed Sensing (VCS) aims to reconstruct multiple frames from one single captured measurement, thus achieving high-speed scene recording with a low-frame-rate sensor. Although there have been impressive advances in VCS recently, those state-of-the-art (SOTA) methods also significantly increase model complexity and suffer from poor generality and robustness, which means that those networks need to be retrained to accommodate the new system. Such limitations hinder the real-time imaging and practical deployment of models. In this work, we propose a Sampling-Priors-Augmented Deep Unfolding Network (SPA-DUN) for efficient and robust VCS reconstruction. Under the optimization-inspired deep unfolding framework, a lightweight and efficient U-net is exploited to downsize the model while improving overall performance. Moreover, the prior knowledge from the sampling model is utilized to dynamically modulate the network features to enable single SPA-DUN to handle arbitrary sampling settings, augmenting interpretability and generality. Extensive experiments on both simulation and real datasets demonstrate that SPA-DUN is not only applicable for various sampling settings with one single model but also achieves SOTA performance with incredible efficiency.
\end{abstract}

%Specifically, we adopted a lightweight U-net as the prior-regularized term in the ADMM algorithm, thereby constructing a DUN that is efficient, optimization-inspired, and E2E trainable. Moreover, we develop a novel Sampling Priors Augmented Learning strategy that feeds the priors from sampling model into DUN to dynamically modulate network features, making the overall network robust to unseen situations.

\keywords{video compressive sensing, computational imaging, deep unfolding network, efficient neural network}

\begin{teaserfigure}
  \centering
  \includegraphics[width=0.9\textwidth]{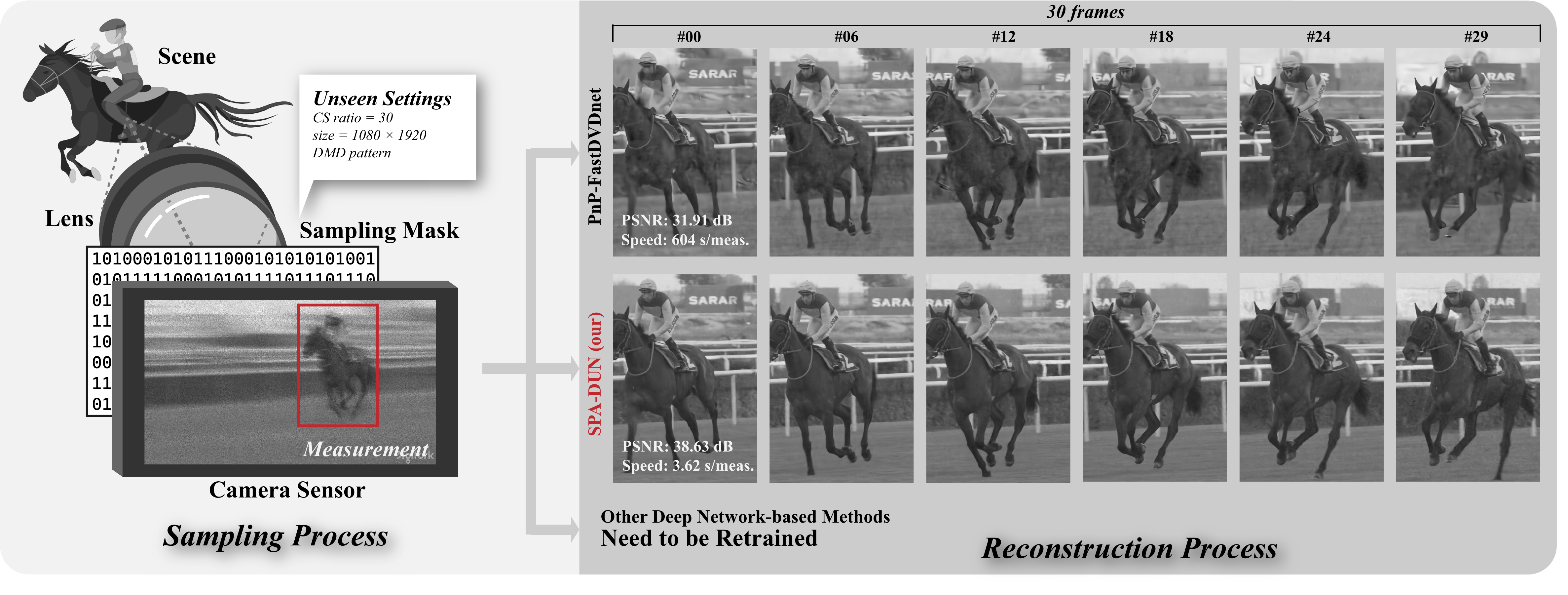}
  \caption{Overview of the VCS system.  The camera sensor encodes multiple frames of the scene through dynamic sampling mask. Our SPA-DUN realizes high-quality reconstruction for unseen sampling settings with one single trained model.}
  \label{fig:camera}
\end{teaserfigure}

\maketitle

\section{Introduction}
As an important branch of computational imaging, inspired by compressive sensing (CS) theory, video compressive sensing (VCS) systems \cite{yuan2021snapshot, reddy2011p2c2, llull2013coded, yuan2014low, qiao2020deep} compress multiple frames along the time dimension into one measurement within a single exposure as shown in Fig.~\ref{fig:camera}. And then, we input the captured measurement and the given sampling mask into a reconstruction algorithm to restore multiple high-quality frames. In this way, a low-frame-rate sensor can achieve ultrafast photography, enjoying the advantages of low-bandwidth, low-power, and low-cost.

Traditional model-based methods regard VCS reconstruction as an optimization problem with image or video prior knowledge as the regularized term. These methods focus on exploiting a structural prior with theoretical guarantees and generalizability, such as sparsity in some transformation domains \cite{yuan2016generalized}, low rank \cite{liu2018rank}, and so on \cite{yang2014compressive, yang2014video}. Although these model-based methods can handle with different scale factors, CS ratios, and mask patterns, the main drawback is that they require manual parameter tuning, which leads to poor generality and slow reconstruction speed.

Over the past few years, deep network-based methods \cite{qiao2020deep, meng2020gap, cheng2020birnat, yang2022ensemble} have accelerated VCS reconstruction and significantly improved the imaging effect by direct learning a nonlinear mapping from the measurements to the original signals. However, most deep network-based methods neglect the VCS problem context. Many advanced but complex designs (eg. 3D convoluton \cite{ji20123d}, Vision Transformer \cite{dosovitskiy2020image}) from general vision have been introduced as a video-to-video network with stronger representation ability. While these advanced designs effectively improve reconstruction performance, they also entail higher training and inference costs. Not only that, these deep network-based methods suffer from poor generality and robustness. These networks were trained for a fixed sampling setting and fail to handle other unseen situations. In real applications, not only the recording target is complex and variable, but also the camera parameters are frequently adjusted for various needs. Therefore, the setting of the sampling system also varies in terms of imaging resolution, CS ratio, and sampling mask pattern. As shown in Fig.~\ref{fig:camera}, most deep network-based methods need to be retrained to accommodate such sampling settings that have not been seen in their training. Obviously, such practices result in large storage space and expensive time costs. Although the model-based methods does not require training, its iterative process is time-consuming, for example, the PnP algorithm \cite{yuan2021plug} takes 604s to reconstruct 30 frames with poor results. Recently, ELP-Unfolding \cite{yang2022ensemble} proposed scalable learning to improve the generality of the model, but the fixed maximum frame of 24 limits further extension.

To address the above issues, we proposed a Sampling-Priors-Augmented Deep Unfolding Network (SPA-DUN) to realize efficient video compressive sensing for arbitrary sampling settings. In order to improve the efficiency of the reconstruction model, we have extracted key components from advanced image-to-image networks \cite{zamir2022restormer,chen2022simple,liang2021swinir,mehta2021evrnet} to obtain a more concise and effective U-net. Based on this lightweight U-net, we unfold the alternating direction multiplier method (ADMM) \cite{boyd2011distributed} to form an end-to-end deep unfolding network (DUN), which enjoys high interpretability and efficiency. To improve the generality, we propose Sampling Priors Augmented Learning (SPA-Learning) strategies, both on the training level and the architectural level. Without resorting to external datasets, we augment the common dataset by random sampling. Besides, our reflective padding enables 2D CNN to be flexible with videos of any lengths while mitigating the counter-impact on the network fitting. And last, the prior knowledge from sampling model are fed into the DUN as explicit physical guidance. In this way, SPA-DUN is able to dynamically modulate the network features for adopting different sampling settings. The major contributions are summarized as follows:
\begin{itemize}
\item We design a lightweight and efficient U-net as the backbone network, which significantly reduces the complexity and increases the capacity of the network.
\item We propose sampling-priors-augmented learning which is exploited to make network robust to unseen sampling settings without retraining.
\item Our SPA-DUN establishes new SOTA in terms of reconstruction effect, model complexity, calculation speed, and generality, promoting the application in real-world VCS systems.
\end{itemize}

\section{Related Work}
\subsection{Video Compressive Sensing}
Video Compressive Sensing is also known as Video Snapshot Compressive Imaging\cite{yuan2021snapshot}, which can be mathematically defined as an ill-posed inverse problem for large-scale linear sampling equation. Traditional model-based approaches treat this ill-posed problem as an optimization problem with a prior-regularized term, such as sparsity in some transformation domains \cite{yuan2016generalized}, low rank \cite{liu2018rank}, and so on \cite{yang2014compressive, yang2014video}. However, these model-based methods not only require iterative solving of optimization problems, but also require manual tuning of different samples, and thus suffer from limited representing capacity, higher latency, and poor generalization ability.

Recently, inspired by the great success of deep learning in image restoration \cite{liang2021swinir,zamir2022restormer,wang2019edvr}, many deep network-based methods have been introduced for accelerating VCS reconstruction. Deep network-based methods directly design E2E networks to learn a nonlinear mapping from the measurement domain to the original signal domain, and then provide instantaneous reconstruction. For example, BIRNAT \cite{cheng2020birnat} employs bidirectional recurrent neural networks to aggregate information from time series. RevSCI \cite{cheng2021memory} adopts reversible 3D convolution to achieve better reconstruction with lower memory consumption. However, the performance of such E2E networks with black-box property is heavily dependent on well-designed architectures. This fact not only results in their tricky training schemes but also drags down their performance, due to the large difficulty of learning recovery mapping without explicit physical guidance.

For explicit physical guidance, Plug-and-Play algorithms \cite{yuan2021plug, yuan2020plug} alternate between minimizing a data-fidelity term to promote data consistency and imposing a learned regularizer in the form of an image denoiser \cite{zhang2018ffdnet, tassano2020fastdvdnet}. This paradigm combines deep networks and interpretable model-based methods to provide flexible and powerful algorithms, but still involve a time-consuming iterative solution process and depend on careful tuning of hyperparameters.  

\subsection{Deep Unfolding Network}
As the main part of physical-inspired CS reconstruction approaches, Deep Unfolding Networks (DUN) \cite{monga2021algorithm} have shown promising performance in many tasks \cite{zhang2020amp,zhang2020deep, zhang2018ista} and usually serve as a key principle for structure design. In the last few years, various DUNs like GAP-net \cite{meng2020gap}, Tensor-FISTA \cite{han2020tensor}, Tensor-ADMM \cite{ma2019deep}, and DUN-3D \cite{wu2021dense} have emerged for VCS reconstruction. The main idea of all of them is to unfold traditional model-based methods into fewer iterations and utilize neural networks to learn partial terms in E2E manner. As the backbone network becomes more advanced, DUN is able to reconstruct more and more details from the measurements. However, previous DUN-based methods have two potential drawbacks: 1) The increasing complexity of the network brings huge training costs and slows down inference. 2) Most previous networks lack generality and robustness. They often suffered significant performance drop or even failed to function at all when sampling settings are changed. Obviously, these two drawbacks hinder the actual deployment and operation of the model. 

Recently, ELP-Unfolding \cite{yang2022ensemble} propose the scalable learning to handle different CS ratios, but is still limited by the fixed maximum frame. The poor generality of DUN is also reported in the field of CS research \cite{zhang2023physics}. COAST \cite{you2021coast} designs a controllable unit to modulate network features by the given hyperparameters, effectively improving the generality of the model. Inspired by this control idea, we extract the prior from the sampling mask and use it to guide the network learning, where such sampling prior is more intuitive and informative for VCS reconstruction.

\begin{figure*}[!t]
\centering
\includegraphics[width=\linewidth]{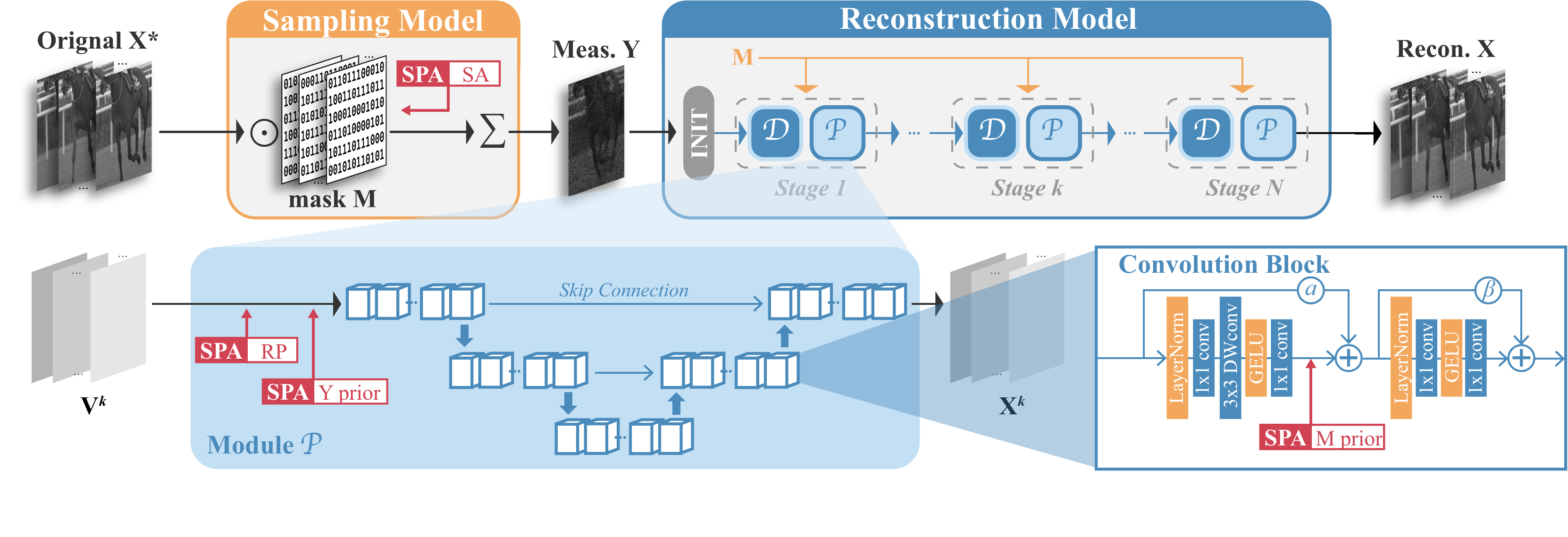}
\caption{Overview of the proposed SPA-DUN, illustrated by (1) Sampling Model (2) Reconstration Model which contains alternating Data-fidelity Modules $\mathcal{D}$ and Prior-regularized Modules $\mathcal{P}$ (3) U-net in Module $\mathcal{P}$ (4) Convolution Block in U-net.}
\label{fig:framework}
\end{figure*}

\section{SPA-DUN}
As shown in Fig.~\ref{fig:framework}, the proposed SPA-DUN is consisted of a sampling model which simulates the capture of the measurements, a reconstruction model which alternates between data-fidelity modules $\mathcal{D}$ and prior-regularized modules $\mathcal{P}$, and several SPA-Learning strategies which enhance generality and robustness. Due to page limitations, we only discuss grayscale imaging problem in the main text, while color imaging problem is given by supplementary material (SM).

\subsection{Sampling Model}
The VCS system consists of a sampling process on the hardware side and a reconstruction process on the algorithm side. During the sampling process, the optical encoder modulates the scene through a given sampling mask $\{\mathbf{M}_t\}_{t=1}^{c} \in \{0,1\}^{h\times w}$ within a single exposure, compressing the image sequence $\{\mathbf{X}_t\}_{t=1}^{c} \in \mathbb{R} ^{h\times w}$ into a 2D measurement $\mathbf{Y} \in \mathbb{R} ^{h\times w}$ along the temporal dimension:
\begin{equation}
\label{eq:SM1}
\mathbf{Y}=\sum_{t=1}^c{\mathbf{M}_t\odot \mathbf{X}_t} +\mathbf{Z}
\end{equation}
where $c$ denotes the CS ratio, $\odot$ denotes the Hadamard (element-wise) product, and $\mathbf{Z} \in \mathbb{R} ^{h\times w}$ is the unknown measurement noise. For easy mathematical description, (\ref{eq:SM1}) is equivalent to the following linear form:
\begin{equation}
\label{eq:SM2}
\boldsymbol{y}=\boldsymbol{\Phi} \boldsymbol{x}+\boldsymbol{z}
\end{equation}
where $\boldsymbol{y}=\operatorname{vec}(\mathbf{Y})\in \mathbb{R} ^{hw}$, $\boldsymbol{x}=\left[\operatorname{vec}(\mathbf{X}_1),\ldots ,\operatorname{vec}(\mathbf{X}_c)\right]\in \mathbb{R} ^{chw}$ , and $\boldsymbol{z}=\operatorname{vec}(\mathbf{Z})\in \mathbb{R} ^{hw}$ are the vectorized representation of tensors $\mathbf{Y}$, $\mathbf{X}$ and $\mathbf{Z}$, respectively. Different from traditional CS problem, the mask $\boldsymbol{\Phi}\in \mathbb{R} ^{hw\times chw}$ in (\ref{eq:SM2}) is a block diagonal matrix consisting of $c$ diagonal matrices shaped as follows:
\begin{equation}
\label{eq:PHI}
\boldsymbol{\Phi} = \left[\mathbf{D}_1, \ldots, \mathbf{D}_c \right]
\end{equation}
where $\mathbf{D}_t=\operatorname{diag}(\operatorname{vec}(\mathbf{M}_t))\in \mathbb{R} ^{hw\times hw}$ for $t = 1,\ldots,c$. The sampling mask is generated by the fully random pattern in the Digtial Micromirror Devices (DMD) \cite{qiao2020deep} or the shifting pattern in the CACTI system \cite{llull2013coded, yuan2014low}. We take only the former (DMD pattern) to build the sampling model in training.

According to this mathematical modeling of the sampling process, we can simulate the capture of measurements. In this way, we can quickly generate sufficient data pairs $(\mathbf{X,Y,M})$ or $(\boldsymbol{x,y,\Phi})$ for training a reconstruction model.

\subsection{Reconstruction Model}
In the following, we will first briefly introduce the ADMM algorithm as preliminary to facilitate the discussion of DUN. Then we will elaborate the details of data-fidelity modules $\mathcal{D}$ and prior-regularized modules $\mathcal{P}$ in proposed SPA-DUN respectively.

\subsubsection{\textbf{DUN based on ADMM}}
From the optimization perspective, the ill-posed inverse problem of solving orignal $\boldsymbol{x}$ in (\ref{eq:SM2}) can be considered as finding the (hopefully unique) $\boldsymbol{x}$ at the intersection of the affine subspace $\boldsymbol{U}=\{\boldsymbol{x}\in \mathbb{R} ^{chw}:\boldsymbol{y}=\boldsymbol{\Phi x}\}$ and the natural video set $\boldsymbol{O}$. It can be formulated as follows:
\begin{equation}
\label{eq:opt1}
\boldsymbol{x}=\arg \min_{\boldsymbol{x}} \frac{1}{2}\|\boldsymbol{y}-\boldsymbol{\Phi x}\|_2^2+\lambda \Psi(\boldsymbol{x})
\end{equation}
where the former data-fidelity term enables $\boldsymbol{x}$ to maintain the consistency of sampling equation, the latter prior-regularized term enables $\boldsymbol{x}$ to match the natural video features, and $\lambda$ balances these two terms. Under the ADMM framework, by introducing an auxiliary vector $\boldsymbol{v}$, the unconstrained optimization in (\ref{eq:opt1}) can be converted into:
\begin{equation}
\label{eq:opt2}
(\hat{\boldsymbol{v}}, \hat{\boldsymbol{x}})=\arg \min _{\boldsymbol{v}, \boldsymbol{x}}\|\boldsymbol{y}-\boldsymbol{\Phi v}\|_2^2+\lambda \Psi(\boldsymbol{x}), \text { s.t. } \boldsymbol{x}=\boldsymbol{v}
\end{equation}
This minimization can be solved by the following sub-problems: 
\begin{subequations}
\begin{align}
\boldsymbol{v}^{(k+1)} & =\arg \min_{\boldsymbol{v}} \frac{1}{2}\|\boldsymbol{y}-\boldsymbol{\Phi v}\|_2^2+\frac{\gamma}{2}\left\|\boldsymbol{v}-\boldsymbol{x}^{(k)}-\boldsymbol{b}^{(k)}\right\|_2^2 \label{eq:opt3A} \\
\boldsymbol{x}^{(k+1)}&=\arg \min_{\boldsymbol{x}} \lambda \Psi(\boldsymbol{x})+\frac{\gamma}{2}\left\|\left(\boldsymbol{v}^{(k+1)}-\boldsymbol{b}^{(k)}\right)-\boldsymbol{x}\right\|_2^2 \label{eq:opt3B}\\
\boldsymbol{b}^{(k+1)}&=\boldsymbol{b}^{(k)}-\left(\boldsymbol{v}^{(k+1)}-\boldsymbol{x}^{(k+1)}\right) \label{eq:opt3C}
\end{align}
\end{subequations}
where $k$ is the number of iterations, and we initialize $\boldsymbol{b}^0=0$, $\boldsymbol{x}^0=\boldsymbol{\Phi}^{\top}\boldsymbol{y}$. 

It can be observed that data-fidelity term and prior-regularized term in (\ref{eq:opt2}) are decoupled to sub-problems (\ref{eq:opt3A}) and (\ref{eq:opt3B}). We unfold these alternating iterative processes into a neural network with $N$ finite stages, where $k$-th iteration of ADMM is cast to $k$-th stage comprising data-fidelity module $\mathcal{D}$ and prior-regularized module $\mathcal{P}$ as shown in Fig.~\ref{fig:framework}.

\subsubsection{\textbf{Data-fidelity Module $\mathcal{D}$}}
Following the above analysis, given $\{\boldsymbol{x, v, \Phi, y}\}$, (\ref{eq:opt3A}) is a quadratic form and has a closed-form solution.
\begin{equation}
\label{eq:d1}
\boldsymbol{v}=\left({\boldsymbol{\Phi}}^{\top} \boldsymbol{\Phi}+\gamma {\mathbf{I}}\right)^{-1}\left[\boldsymbol{\Phi}^{\top} \boldsymbol{y}+\gamma(\boldsymbol{x}+\boldsymbol{b})\right]
\end{equation}
Due to the special structure of $\boldsymbol{\Phi}$, $\boldsymbol{\Phi\Phi^{\top}}$ is a diagonal matrix and can be defined as:
\begin{equation}
\label{eq:d2}
\boldsymbol{\Phi \Phi^{\top} }\stackrel{\operatorname{def}}{=} \operatorname{diag}\left\{\psi_1, \ldots, \psi_{hw}\right\}
\end{equation}
As proved in DeSCI \cite{liu2018rank}, (\ref{eq:d2}) can be solved in one shot:
\begin{equation}
\label{eq:d3}
\boldsymbol{v} = (\boldsymbol{x}+\boldsymbol{b})+ 
\boldsymbol{\Phi}^{\top}\left[\frac{\boldsymbol{y}_1-[\boldsymbol{\Phi}(\boldsymbol{x}+\boldsymbol{b})]_1}{\gamma+\psi_1}, \ldots, \frac{\boldsymbol{y}_{hw}-[\boldsymbol{\Phi}(\boldsymbol{x}+\boldsymbol{b})]_{hw}}{\gamma+\psi_{hw}}\right]^{\top}
\end{equation}

After this projection, $\boldsymbol{v}$ (or tensor $\mathbf{V}$) will be close to the fidelity domain, i.e., guaranteeing the consistency of the sampling equation in (\ref{eq:SM2}). Moreover, we set the penalty coefficient $\gamma$ as a learnable parameter to enhance the flexibility of the reconstruction process.

\subsubsection{\textbf{Prior-regularized Module $\mathcal{P}$}}
\label{sec:moduleP}
For prior-regularized term, it is difficult to define a mathematically feasible and practically effective constraint $\Psi(\cdot)$ with natural video features and derive a closed-form solution. Therefore, similar to previous DUN methods, we employ a deep network $\textsc{Net}_\theta(\cdot)$ which maps from degraded video to high-quality video to replace $\Psi(\cdot)$. In other words, the network will learn prior knowledge from numerous training data, thus acting as a regularization of (\ref{eq:opt3B}) in ADMM. 
\begin{equation}
\label{eq:p1}
\mathcal{P}: \mathbf{X}=\textsc{Net}_\theta(\mathbf{V-B})
\end{equation}

Previous works usually employ a more advanced and complex video-to-video network to improve the representation ability. However, the paradigm of DUN, which sequentially stacks multiple networks, inevitably magnifies the overall complexity and drags down the inference speed. To realize the trade-off between the model's computational cost and quality, we design a lightweight U-net as the prior-regularized module $\mathcal{P}$. This U-net contains MLPMixer-inspired convolution blocks as shown in the lower right of Fig.~\ref{fig:framework}. 

In details, we utilize depthwise (DW) convolution \cite{howard2017mobilenets} and $1\times1$ convolution as a combination. This popular combination not only drastically reduces the complexity compared to the native convolution, but also improves the performance of the network on many other vision tasks \cite{chen2022simple,mehta2021evrnet,yu2021lite} by increasing the cardinality \cite{xie2017aggregated} of the features. Inspired by MLPMixer \cite{tolstikhin2021mlp}, we add two residual connections with learnable scaling factors to form a spatial mixer and a channel mixer. Besides, we retain GELU \cite{hendrycks2016gaussian} and LayerNorm \cite{ba2016layer}, which are common in Transformer and also work in CNN \cite{liu2022convnet}. In section~\ref{sec:ab_unet}, we implement several U-nets with different types of blocks for comparison, which shows that our MLPMixer-inspired design is efficient for such low-semantic video-to-video mapping.

\subsection{Sampling Priors Augmented Learning}
To realize generality and robustness for unseen sampling settings, we propose novel Sampling Priors Augmented Learning strategies, both at the training level and the architectural level.

\subsubsection{\textbf{Sampling Augmentation (SA)}}
\begin{figure}
\centering
\includegraphics[width=0.8\linewidth]{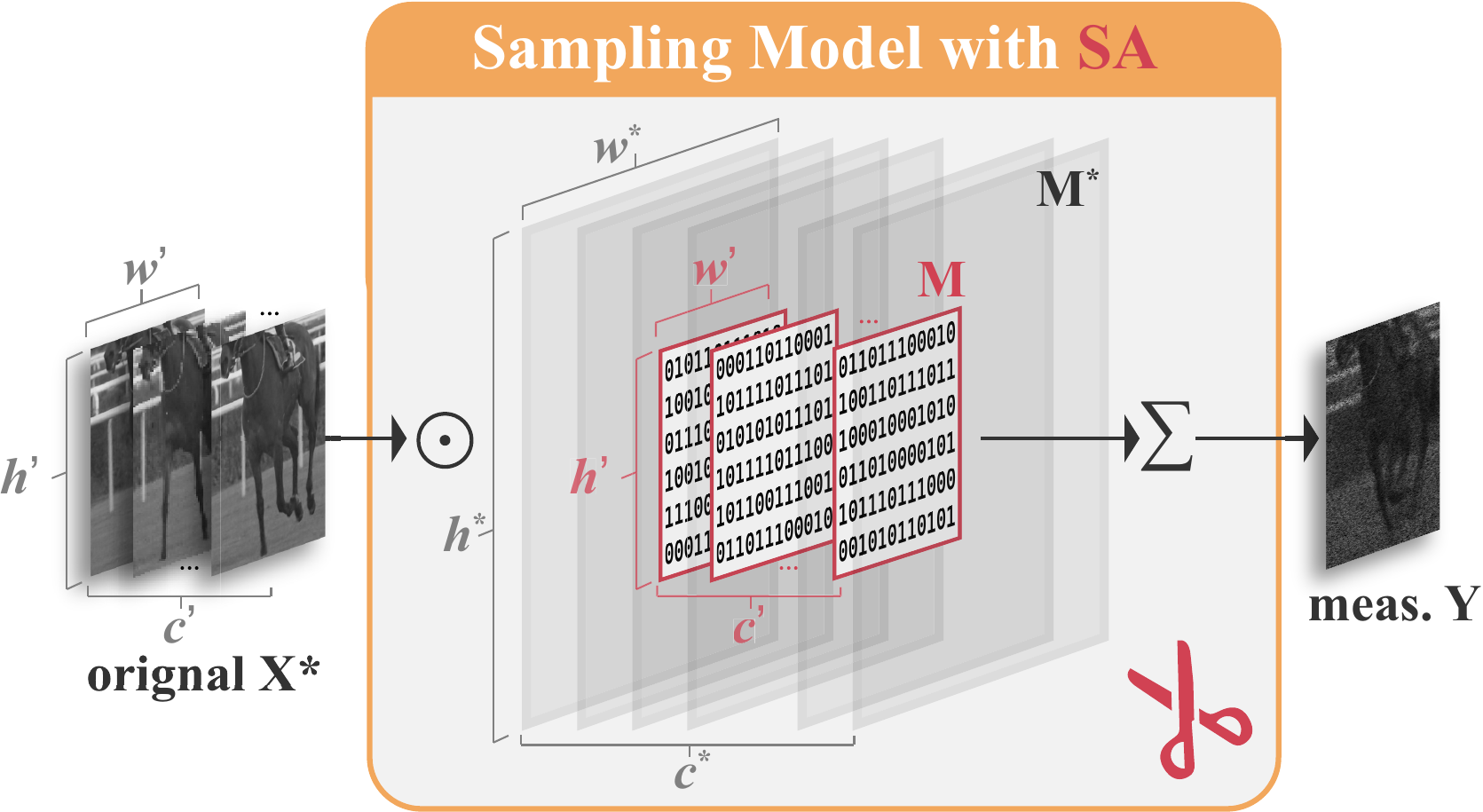}
\caption{The idea of the Sampling Augmentation (SA) strategy. Randomly cropping the mask to augment the sampling model in training.}
\label{fig:RSA}
\end{figure}

The proposed SA is only adopted at the training stage of sampling model as shown in the Fig.~\ref{fig:RSA}. Given a selection set of CS ratios $\boldsymbol{S}=\{c_i\}_{i=0}^{n}$ and a sampling mask $\mathbf{M}^* \in \{0,1\}^{c^* \times h^*\times w^*}$ with sufficient size, we randomly crop out a patch $\mathbf{M} \in \{0,1\}^{c'\times h'\times w'}$ where $c'\in \mathbf{S}$, and then generate the corresponding measurements in each small batch of training. 

As a result, the SA strategy promotes the training diversity by cropping out various sampling settings from one fixed mask. This low-cost strategy can alleviate the overfitting problem of network similar to the regular data augmentation techniques. Meanwhile, the learning from different sampling settings will significantly improve the generalization capability. The effectiveness of SA will be validated in section~\ref{sec:ab_spa}.

\subsubsection{\textbf{Reflective Padding (RP)}}
\label{sec:rp}
Although the module $\mathcal{P}$ in our DUN is fully convolution network that can input sequences with any spatial sizes, it cannot function on sequences with different CS ratios (i.e., temporal sizes) due to the inherent limitations of 2D convolution. ELP-Unfolding \cite{yang2022ensemble} fixed the temporal size of the input to a maximum value $L$, and padded the data less than $L$ frames by repetitive arrangement. In this work, we upgrade this simple padding to reflective padding (RP) as:
\begin{equation}
\label{eq:rp}
\resizebox{\linewidth}{!}{$
\textsc{RP}(A)=\left\{
\begin{aligned}
& \operatorname{cat}[\{\mathbf{A}_1\dots \mathbf{A}_c\},\{\mathbf{A}_c\dots \mathbf{A}_1\},\dots]_1^{[0:L]} & c < L\\
& \operatorname{cat}[\{\mathbf{A}_1\dots \mathbf{A}_L\},\{\mathbf{A}_{L+1}\dots \mathbf{A}_{2L}\},\dots,\{\mathbf{A}_{c-L+1}\dots \mathbf{A}_{c}\}]_0  & c \geq L\\
\end{aligned}
\right.$}
\end{equation}
where $\operatorname{cat}[ ]_0$ and $\operatorname{cat}[ ]_1$ denotes the concatenation along the batch dimension and temporal dimension, respectively. For an image sequence (video) $A\in R^{b\times c\times h\times w}$ where b is the batch size, if the temporal size $c<L$, we append its reverse sequence at the end and repeat $T$ times until $Tc>=L$. If $c\geq L$, we input the subsequences with $L$ frames into the network in batches, where the last subsequence less than $L$ frames will be backfilled into $L$ frames. 

In this way, the output sequences $(\mathbf{V-B})$ with various $c$ from the former module $\mathcal{D}$ are padded into $\textsc{RP}(\mathbf{V-B})$ of shape $[b',L,h,w]$, where $b'=b\times\textsc{Roundup}(c/L)$, and are fed into the 2D CNN in module $\mathcal{P}$. Compared to the previous simple padding, this low-cost reflective padding not only makes the 2D CNN flexible for arbitrary inputs without upper limit, but also has smoother inter-frame transitions to reduce the difficulty of network learning. 

\subsubsection{\textbf{Sampling Priors (SP)}} 
\begin{figure}
\centering
\includegraphics[width=\linewidth]{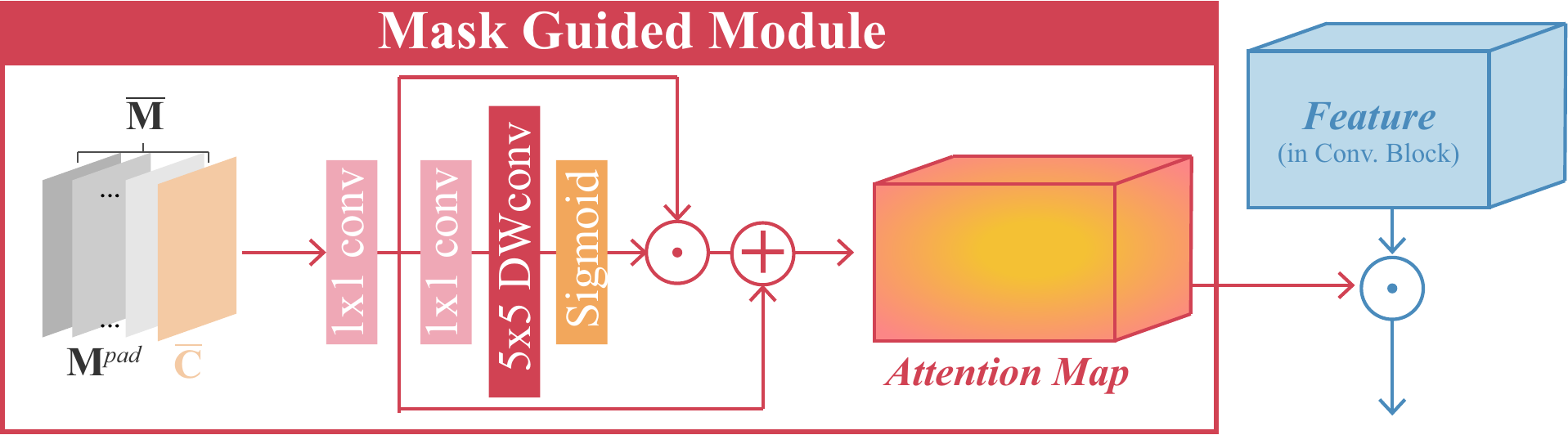}
\caption{The detailed structure of the Mask Guided Module.}
\label{fig:MGM}
\end{figure}

If the module $\mathcal{P}$ takes only the fidelity output $(\mathbf{V-B})$ as input, it may not be able to sense and adapt the changes in sampling model. To compensate for missing information, we extract and feed the priors of sampling model to the module $\mathcal{P}$. Specifically, we first normalize measurements by $\overline{\mathbf{Y}} = \mathbf{Y} \oslash \sum_{t=1}^{c}{\mathbf{M}_t}$. Since the normalized $\overline{\mathbf{Y}}$ is closer to the fidelity output $(\mathbf{V-B})$ in distribution, we can concatenate these two as:
\begin{equation}
\overline{\mathbf{V}} = \operatorname{cat}[\operatorname{RP}(\mathbf{V-B}),\overline{\mathbf{Y}}]_1 \label{eq:ycat}
\end{equation}
And then, we use $\overline{\mathbf{V}}$ as the first layer input of the network in the module $\mathcal{P}$. Moreover, a lightweight Mask Guided Module \cite{cai2022mask} is introduced to sense changes in the sampling mask and further modulate the network features as shown in Fig.~\ref{fig:MGM}. The input of this module consists of the following concatenation:
\begin{equation}
\overline{\mathbf{M}} = \operatorname{cat}[\operatorname{RP}(\mathbf{M}), \overline{\mathbf{C}}]_1 = \operatorname{cat}[\operatorname{RP}(\mathbf{M}),\operatorname{span}(c'/L)]_1
\label{eq:minput}
\end{equation}
where the operation $\operatorname{span}(\cdot)$ duplicates the constant $c'/L$ into a 2D matrix $\overline{\mathbf{C}}$, replenishing the missing length information. After passing through several $1\times1$ convolutions and $5\times5$ DW convolutions, we use the output attention maps to modulate the stem features in the convolution blocks. 

In this way, the priors from the measurements and sampling masks are exploited to augment the network in a reasonable way. On the one hand, those extra priors can be regarded as physical guidance to reduce the difficulty of learning recovery mapping. On the other hand, when the sampling model changes, the network can directly sense these changes and dynamically modulate the features. 

\begin{table*}
  \caption{Average PSNR/SSIM performance comparisons on six grayscale benchmark datasets with various sampling settings. Seen or unseen depends on whether it is the sampling mask used for training. The best results are highlighted in bold text. \label{tab:scale}}
  \centering
  \resizebox{\linewidth}{!}{
  \begin{tabular}{clcccccccc}
    \toprule
    \multirow{2}{*}{Type} & \multirow{2}{*}{Method} & \multirow{2}{*}{Pattern} & \multicolumn{7}{c}{PSNR(dB), SSIM} \\
    \cmidrule{4-10}
    & & & c=8 & c=12 & c=16 & c=20 & c=24 & c=28 & c=32 \\
    \midrule
    \multirow{5}{*}{Seen} & PnP-FastDVDnet \cite{yuan2021plug} & DMD & 32.27, 0.9346 & 30.73, 0.9112 & 29.46, 0.8851&28.63, 0.8625&27.96, 0.8410&27.14, 0.8145&26.38, 0.7896\\
    & RevSCI \cite{cheng2021memory} & CACTI & 33.81, 0.9566&26.48, 0.8611&23.04, 0.7622&21.55, 0.6990&20.86, 0.6653&20.25, 0.6352&19.85, 0.6135\\
    & DUN-3D \cite{wu2021dense} & CACTI & 35.28, 0.9678&32.97, 0.9516&28.59, 0.9021&25.16, 0.8303&23.27, 0.7706&22.01, 0.7150&21.15, 0.6679\\
    & ELP-Unfolding \cite{yang2022ensemble} & DMD & 34.54, 0.9640&33.22, 0.9507&32.08, 0.9363&31.40, 0.9259&26.21, 0.7546&\multicolumn{2}{c}{Not Supported}\\
    & \textbf{SPA-DUN} & DMD & \textbf{35.46, 0.9697}&\textbf{33.47, 0.9510}&\textbf{32.35, 0.9381}&\textbf{31.65, 0.9272}&\textbf{31.38, 0.9218}&\textbf{30.35, 0.9043}&\textbf{28.94, 0.8784}\\
    \midrule
    \multirow{5}{*}{Unseen} & PnP-FastDVDnet \cite{yuan2021plug} & CACTI & 31.90, 0.9298&30.07, 0.9048&28.55, 0.8750&27.41, 0.8406&26.43, 0.8077&25.44, 0.7694&24.59, 0.7392\\
    & RevSCI \cite{cheng2021memory} & DMD & 17.68, 0.5084&17.54, 0.4702&17.17, 0.4333&16.93, 0.4113&16.79, 0.4008&16.64, 0.3870&16.52, 0.3759\\ 
    & DUN-3D \cite{wu2021dense} & DMD & 31.51, 0.9334&28.19, 0.8845&24.32, 0.7888&21.82, 0.6981&20.39, 0.6336&19.33, 0.5770&18.51, 0.5263\\
    & ELP-Unfolding \cite{yang2022ensemble} & CACTI &33.71, 0.9599&31.62, 0.9411&29.20, 0.9102&27.12, 0.8753&23.02, 0.6647 & \multicolumn{2}{c}{Not Supported}\\
    & \textbf{SPA-DUN} & CACTI &\textbf{34.94, 0.9672}&\textbf{32.56, 0.9461}&\textbf{30.76, 0.9250}&\textbf{29.30, 0.9037}&\textbf{28.33, 0.8886}&\textbf{26.34, 0.8485}&\textbf{24.72, 0.8059}\\
    \bottomrule
  \end{tabular}}
\end{table*}

\section{Experiments}
\subsection{Experimental Settings}
\subsubsection{\textbf{Datasets}}
Following previous research \cite{yang2022ensemble,wu2021dense}, we selected 150 scenes at 480p resolution from the DAVIS2017 dataset \cite{pont20172017} as our training dataset. We cropped the original frames into $128\times 128$ patches to reduce training burden. According to the sampling model and SA strategy, we can simulate the sampling process to generate measurements for training. 

To evaluate the basic performance of the model, we utilized six grayscale benchmark datasets including {\tt{Areial, Crash, Drop, Kobe, Runner}}, and {\tt{Traffic}} with a size of $256\times 256$, following the setup in \cite{yuan2021plug}. To assess the generality of the model, we added four large-scale datasets\cite{mercat2020uvg} including {\tt{Beauty, Bosphorus, Jockey}}, and {\tt{ShakeNDry}}, with a size of $1080\times 1920$.

\subsubsection{\textbf{Implementation Details}}
SPA-DUN uses the same U-net design for each module $\mathcal{P}$. Specifically, each U-net has 4, 6, and 4 convolution blocks respectively at three scales. The channel width of the first scale is set to 48 and is doubled after every downsampling layer. To achieve a better trade-off, the default stage number $N$ is set to 10. For SPA-Learning, we set $L=24$ and $\mathbf{S}=\{8,14,18,24\}$. Lastly, the loss function is designed to the weighted RMSE between the ground truth $\mathbf{X}^*$ and the reconstructed outputs $\mathbf{X}^N, \mathbf{X}^{N-1}, \mathbf{X}^{N-2}$ from the last three stages as: 
\begin{equation}
\label{eq:loss}
\mathcal{L}(\theta) = {\sqrt{||\mathbf{X}^*-\mathbf{X}^N||_2^2}}+0.5{\sqrt{||\mathbf{X}^*-\mathbf{X}^{N-1}||_2^2}} 
+0.5{\sqrt{||\mathbf{X}^*-\mathbf{X}^{N-2}||_2^2}}
\end{equation}

We trained SPA-DUN using AdamW optimization \cite{loshchilov2017decoupled} with a batch size of 6. During the first 1000 epochs, we set the learning rate to 1e-3 for faster convergence. In the next 5000 epochs, the learning rate was decayed by 90\% every 300 epochs to reduce oscillation. The training of SPA-DUN lasted for roughly six A100 days.

\begin{figure*}
\centering
\includegraphics[width=0.85\linewidth]{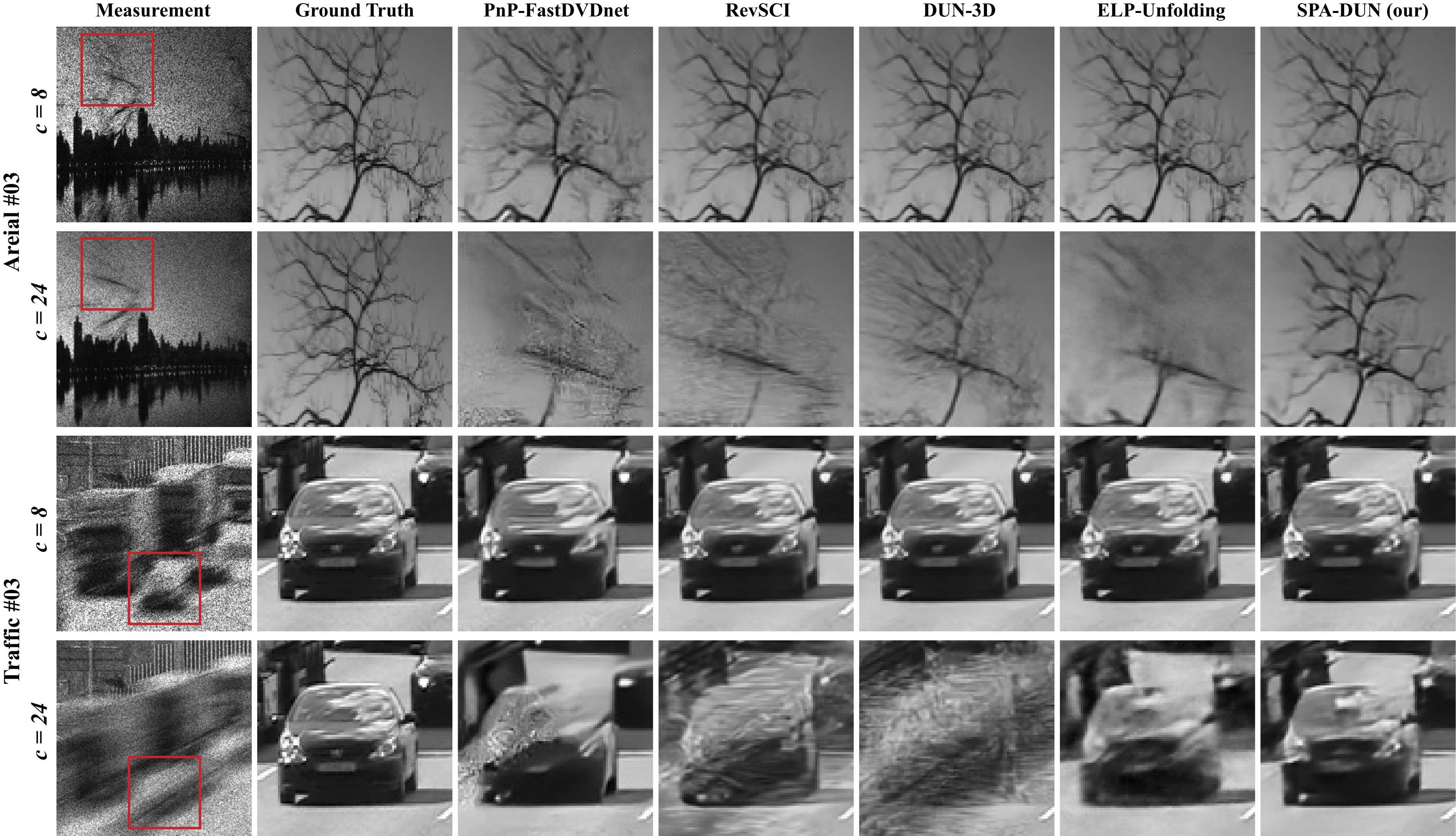}
\caption{Visual comparison on benchmark datasets under seen mask pattern in the case of c=8 and c=24. \#03 indicates the 3rd frame. Full videos are provided in SM.}
\label{fig:256}
\end{figure*}

\subsection{Comparison with State-of-the-Art Methods}
\subsubsection{\textbf{Benchmark Datasets}}
We compared our proposed SPA-DUN with recent representative methods, including PnP \cite{yuan2021plug}, RevSCI \cite{cheng2021memory}, DUN-3D \cite{wu2021dense}, and ELP-Unfolding \cite{yang2022ensemble}. The average PSNR/SSIM performance on six grayscale benchmark datasets with different sampling settings are summarized in Table~\ref{tab:scale}. "Seen" means that the testing mask pattern is the same as the training mask pattern. "Unseen" means that if a method used DMD pattern during training, we changed it to CACTI pattern during testing and vice versa. It's worth noting that all deep network-based methods in this comparison are trained by the same training datasets and are validated by one single trained model without any extra fine-tuning or retraining. 

Table~\ref{tab:scale} shows that SPA-DUN outperforms significantly other methods at all CS ratios, both for seen and unseen mask patterns, benefiting from the proposed SPA-Learning. We displayed some selected reconstructed results under seen mask pattern in Fig.~\ref{fig:256}. SPA-DUN is able to recover more details of high-speed moving objects (branches and vehicles) under extreme conditions ($c=24$), while the reconstructions of other methods have been highly distorted. 

\begin{figure}
  \centering
  \includegraphics[width=\linewidth]{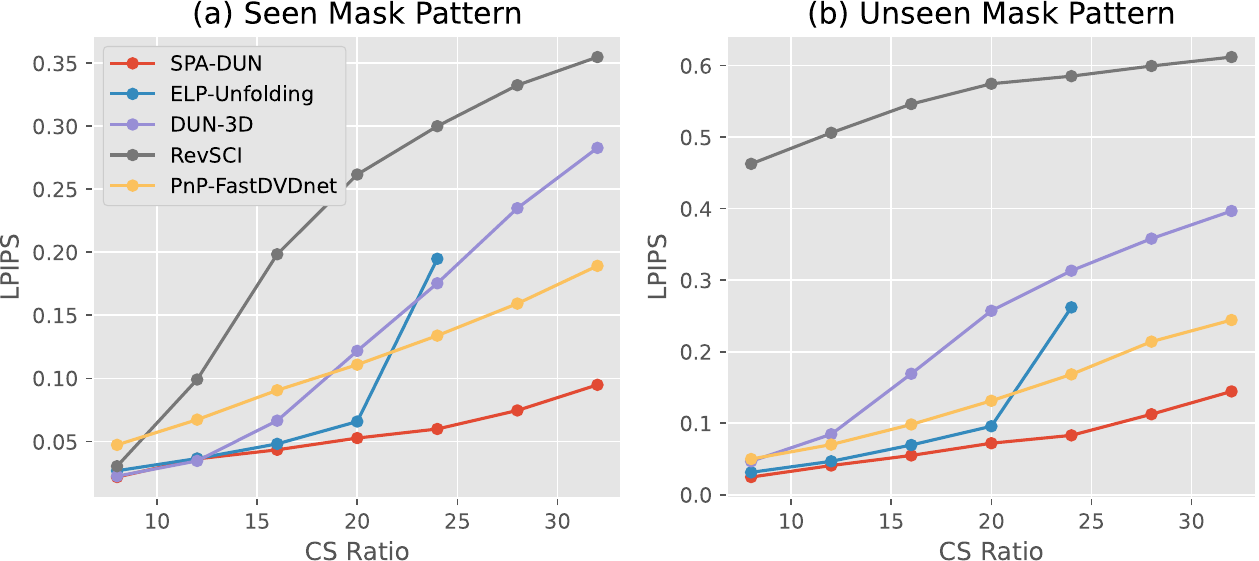}
  \caption{Average LPIPS curves on the benchmark datasets with various CS ratios under seen/unseen mask pattern.}
  \label{fig:multicr}
  \end{figure}

We also ploted intuitive performance curves in Fig.~\ref{fig:multicr} (a) and (b), where LPIPS \cite{zhang2018unreasonable} (lower value indicates better performance) is closer to human perception and suitable for evaluating these highly distorted results. Compared to ELP-Unfolding with scalable learning, SPA-DUN is not limited by the maximum frame and leads significantly at high CS ratios. In terms of performance degradation, the downtrend of SPA-DUN is even flatter than the PnP method which iteratively solves for each sample, demonstrating excellent robustness.

\begin{table*}
  \caption{Average PSNR/SSIM performance comparisons on four large-scale datasets under seen mask pattern at c=24. These metrics are counted in the same hardware environment (A100-80GiB). \label{tab:1080}}
  \centering
  \resizebox{\linewidth}{!}{
  \begin{tabular}{lccccccccc}
    \toprule
    \multirow{2}{*}{Method} & Params & FLOPs & Speed & GPU MEM & \multicolumn{5}{c}{ PSNR(dB), SSIM } \\
    \cmidrule{6-10}
    & (M) & (T) & (s/meas.) & (GiB) & Beauty & Bosphorus & Jockey& ShakeNDry & Average \\
    \midrule
    PnP-FFDnet & -& -& 253.32 & 9.46 & 35.00, 0.8515& 32.05, 0.8586&34.57, 0.8586&26.39, 0.6997 & 32.01, 0.8171\\
    PnP-FastDVDnet & -& -& 395.62 & \textbf{5.17} & 32.40, 0.7591&34.37, 0.8838&32.62, 0.8765&31.97, 0.8380&32.84, 0.8393\\
    RevSCI & \textbf{5.66} & 72.66 & 5.65 & 39.29 & 1.83, 0.3147 & 7.07, 0.4052 & 2.79, 0.3369 & 6.34, 0.3380 & 4.512, 0.3487\\
    DUN-3D & 61.91 & \multicolumn{3}{c}{Out of Memory}\\
    ELP-Unfolding & 567.15 & 149.59 & 3.91 & 47.06 & 30.15, 0.7152&33.93, 0.8802&30.95, 0.8171&30.08, 0.7993&31.28, 0.8029\\
    \textbf{SPA-DUN} & 41.21 & \textbf{17.15} & \textbf{1.21} & 12.47& \textbf{38.29, 0.8951}&\textbf{40.51, 0.9638}&\textbf{38.63, 0.9405}&\textbf{35.43, 0.9081}&\textbf{38.21, 0.9269}\\
    \bottomrule
  \end{tabular}}
\end{table*}

\begin{figure*}
\centering
\includegraphics[width=0.85\linewidth]{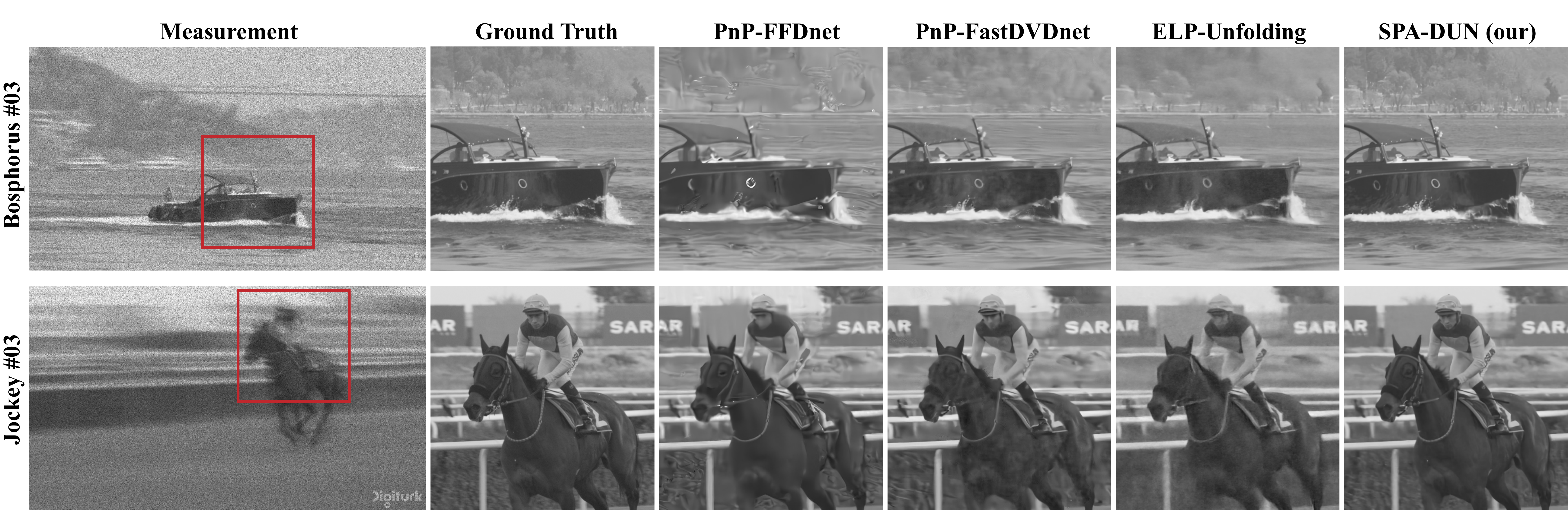}
\caption{Visual comparisons on large-scale datasets under seen mask pattern in the case of c=24. Full videos are provided in SM.}
\label{fig:1080}
\end{figure*}

\subsubsection{\textbf{Large-Scale Datasets}}
To verify the high-resolution imaging capability required for realistic applications, we introduce several large-scale datasets with a size of $1080 \times 1920$ and set the CS ratio to 24. The quantitative results are reported in Table~\ref{tab:1080}. Noted that RevSCI failed to output as expected and DUN-3D is out of GPU memory. On the contrary, our SPA-DUN maintains the same performance superiority as the benchmark datasets. Meanwhile, SPA-DUN leads other approaches by a large margin in terms of model complexity, calculation speed and GPU memory usage, due to the efficient network structure. As shown in the Fig.~\ref{fig:1080}, SPA-DUN can recover more details (waves and human faces), making the VCS process nearly lossless. This advantages promote real-time imaging applications on mobile devices.

\subsection{Ablation Study}
\subsubsection{\textbf{Validating the Efficiency of our U-net}}
\label{sec:ab_unet}
To verify the efficiency of the convolution block proposed in section~\ref{sec:moduleP}, we used the classical ResNet block \cite{he2016deep} and ResNeXt block \cite{xie2017aggregated} as comparisons. We adopted a single U-net to learn the mapping from the measurement to the original signal without unfolding, which provides a more intuitive assessment for the fitting ability of the convolution blocks itself. It is worth noting that our block contains double residual connections and more convolutions, so we reduced the number of blocks to half for a fair comparison. 

\begin{table}[h]
  \caption{Ablation study on the efficiency of our designed U-net. Average PSNR/SSIM at c=8 on benchmark datasets.\label{tab:ab_unet}}
  \centering
  \resizebox{\linewidth}{!}{
  \begin{tabular}{lccccccc}
    \toprule
    Block Type & Num Blocks & Width & Params & FLOPs & PSNR & SSIM \\
    \midrule
    ResNet & 4 4 4 & 48 & 4.51M & 56.36G & 30.13 & 0.903\\
    ResNeXt & 4 4 4 & 48 & 1.20M & 15.32G & 28.97 & 0.877\\
    Our & 2 2 2 & 48 & 1.18M &  14.12G & 31.90 & 0.937\\
    \bottomrule
  \end{tabular}}
\end{table}

\begin{figure}
\centering
\includegraphics[width=\linewidth]{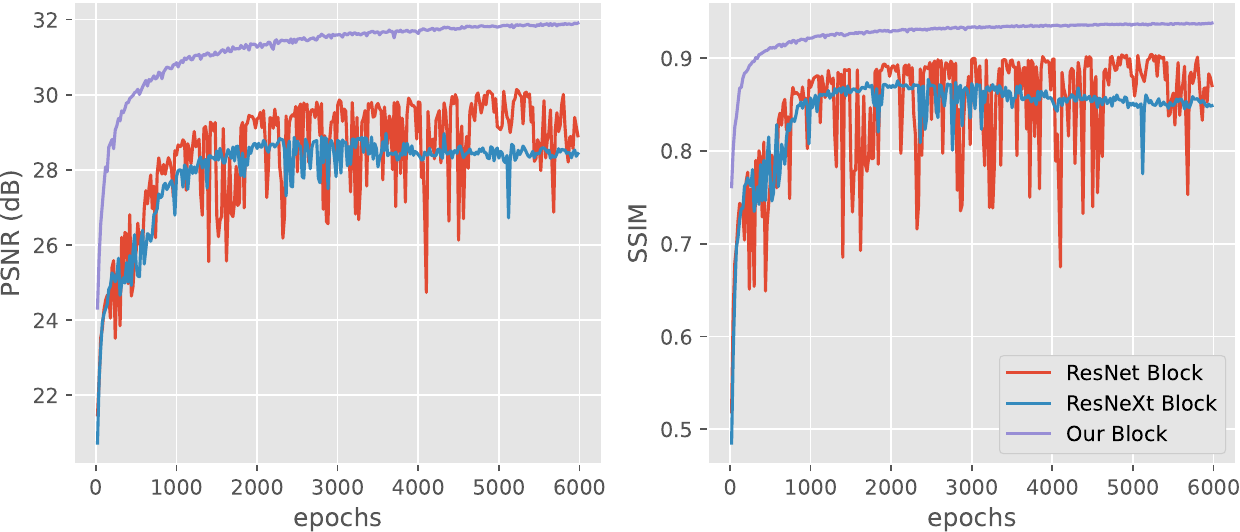}
\caption{Average PSNR/SSIM curves on the benchmark datasets in training, corresponding to Table~\ref{tab:ab_unet}.}
\label{fig:unet}
\end{figure}

We used the benchmark datasets as the validation sets for the training process and record the results in the Fig.~\ref{fig:unet} and Table~\ref{tab:ab_unet}. Compared to ResNet block, ResNeXt block with DW convolution has more stable training and lower model complexity, sacrificing some reconstruction accuracy. Benefiting from the MLPMixer-inspired layer ordering, our lightweight design effectively increases the capacity of the network and thus achieves a significant lead in the grayscale benchmark.

\begin{table*}
  \caption{Ablation study on effects of components in the proposed SPA-Learning, where ReP denotes repetitive padding in ELP, ReF denotes our reflective padding, and CatB denotes the concatenation along the batch dimension in (\ref{eq:rp}) when $c\geq L$. All schemes employ CatB to evaluate the generality for different sampling settings.\label{tab:ab_spa}}
  \centering
  \begin{tabular}{c|c|c|cc|c|c|cc|ccc|cc}
      \toprule
      \multirow{2}{*}{Scheme} & \multirow{2}{*}{SA} & \multirow{2}{*}{Padding} & \multicolumn{2}{c|}{SP} & Training & \multirow{2}{*}{Params} & \multicolumn{2}{|c}{Seen} & \multicolumn{3}{|c}{Unseen CS ratio} & \multicolumn{2}{|c}{Unseen Pattern} \\
      \cmidrule{4-5}
      \cmidrule{8-14}
      & & & Y & M & CS ratios & & c=8 & c=18 & c=10 & c=20 & c=30 & c=8 & c=18 \\
      \midrule
      1 & - & CatB & - & - & \{8\}& 9.47M & 35.40&27.55&31.61&26.33&24.36&33.50&25.39 \\
      2 & \checkmark & CatB+ReP & - & - &\{8,14,18,24\} & 9.54M & 32.30&29.49&29.11&26.56&26.21&31.85&27.86  \\
      3 & \checkmark & CatB+ReF & - & - &\{8,14,18,24\} & 9.54M & 33.05&30.56&30.51&29.47&27.71&32.36&28.18 \\
      4 & \checkmark & CatB+ReF & \checkmark & - &\{8,14,18,24\} & 9.66M & 33.16&30.68&30.79&29.56&27.83&32.50&28.51 \\
      5 & \checkmark & CatB+ReF & \checkmark & \checkmark &\{8,14,18,24\} & 10.88M & 34.38&31.74&32.12&30.53&28.61&33.75&29.47 \\
      \bottomrule
  \end{tabular}
\end{table*}

\subsubsection{\textbf{Validating the Scalability of SPA-DUN}}
\label{sec:ab_spa}
This subsection will present the ablation study to investigate the contribution of each component in our proposed SPA-DUN. To save computing resources, we conducted ablation studies on a shallower SPA-DUN with $N=5$ and $num\_blocks=[2,3,2]$, and reported the average PSNR results on benchmark datasets in Table~\ref{tab:ab_spa}. 

\textbf{Effect of SA} Scheme 1 is a baseline trained by a fixed mask at $c=8$. Scheme 2 is similar to the scalable learning used in ELP-Unfolding, which includes the SA strategy and ReP padding for diverse the sampling settings. The comparison results show that SA enables one single model to be robust for unseen sampling settings, but sacrifices the performance in the specific setting ($c=8$).

\textbf{Effect of RP} Compared to scheme 2, scheme 3 adopted the ReF padding. Such an nearly zero-cost modification can improve the model by 0.32$\sim$2.91 dB overall, especially for the unseen CS ratios. This indicates that such natural inter-frame transition is beneficial for network learning. 

\textbf{Effect of SP} Compared to scheme 3, scheme 4 additionally adopted normalized measurement as the input of module $\mathcal{P}$. The overall performance can be slightly improved by 0.11$\sim$0.33 dB. Scheme 5 further utilized the sampling mask as physical guidance, which allows the network to dynamically adapt to changes in the sampling mask, resulting in significant boosts of about 0.78$\sim$1.33 dB in the seen mask and 0.96$\sim$1.25 dB in the unseen mask. 

\begin{figure}
\centering
\includegraphics[width=0.6\linewidth]{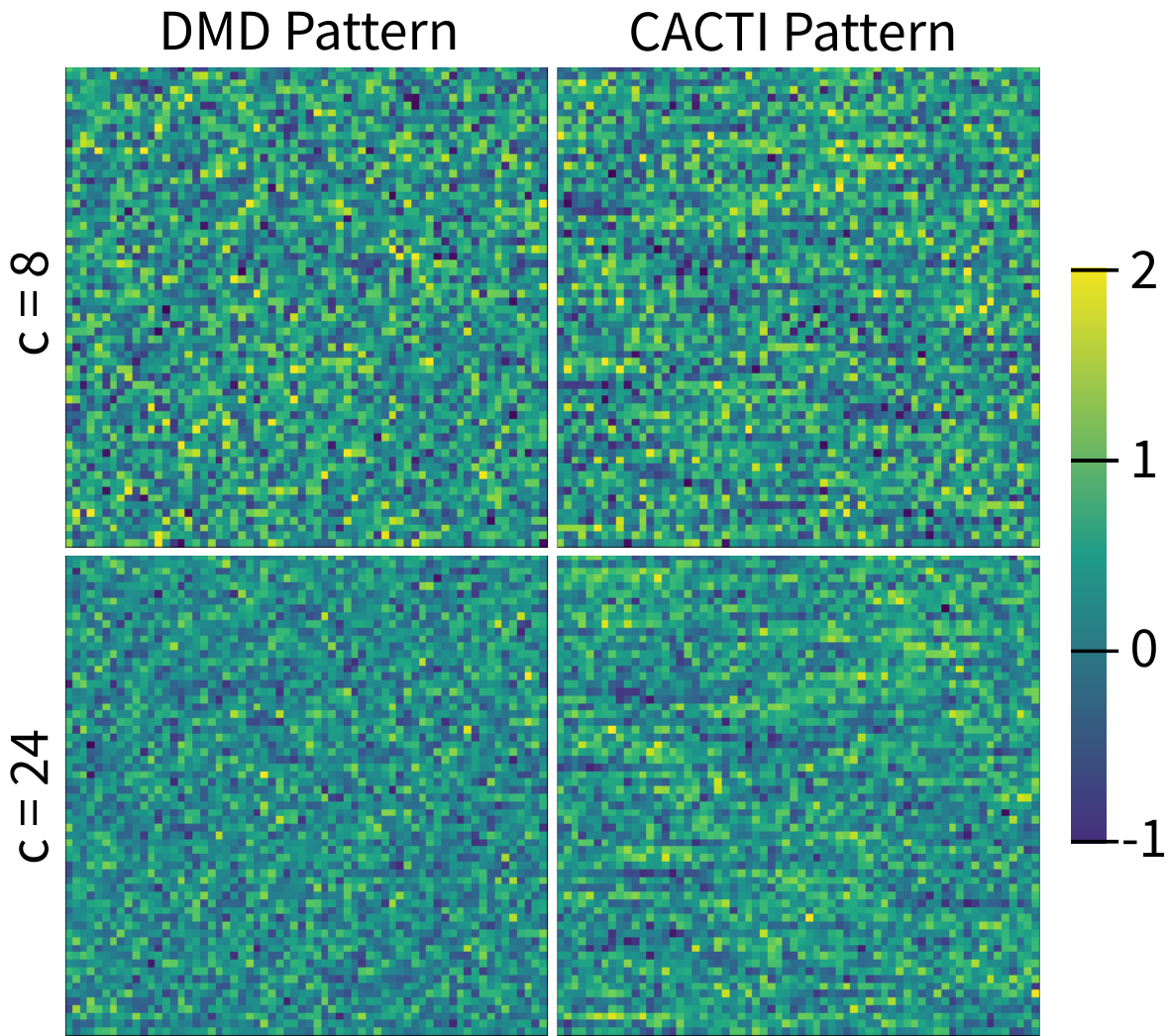}
\caption{Attention visualization in the mask guided module.}
\label{fig:att}
\end{figure}

Furthermore, we visualized the attention in the mask guided module with different sampling settings. As illustrated in Fig.~\ref{fig:att}, the attention map for CACTI pattern displays a clear horizontal stretching texture, which corresponds to the shifting nature behind CACTI pattern. As the CS ratio increases, the horizontal texture in the attention map is further stretched. At the same time, the average value is smaller to ensure the final output energy is stable. We conclude that this mask guided module is able to perceive changes explicitly and then impose the learned attention map on the network features, forming an adaptive paradigm.

\subsection{Real Applications}
\begin{figure}[h]
\centering
\includegraphics[width=\linewidth]{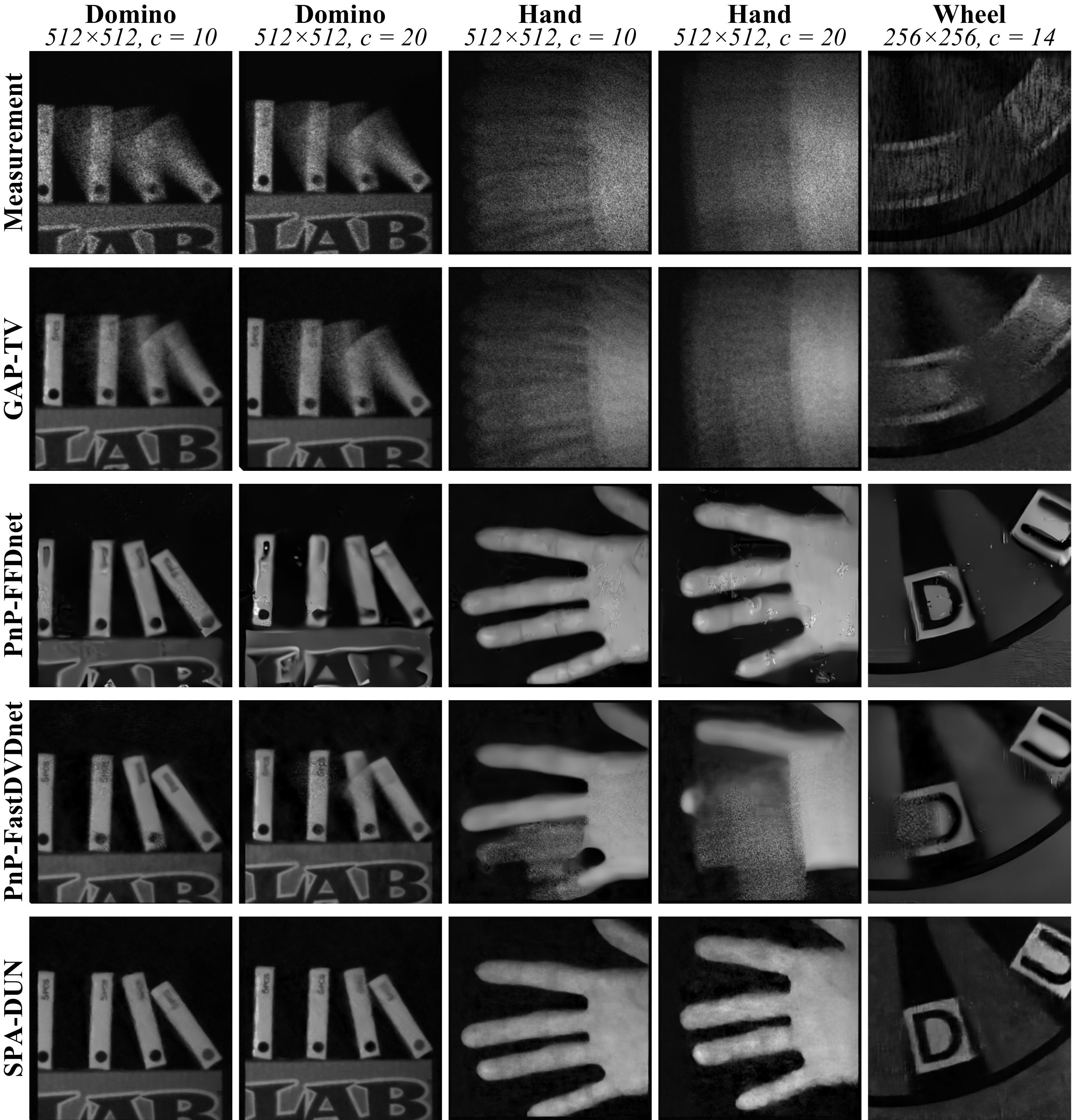}
\caption{Visual comparisons with other available methods on real captured datasets. Full videos are provided in SM.}
\label{fig:real}
\end{figure}

We evaluate SPA-DUN on several real datasets captured by two VCS system \cite{llull2013coded,qiao2020deep}. The {\tt{Domino}} and {\tt{Hand}} data were modulated by DMD \cite{qiao2020deep} with $c=10$ and $c=20$. The {\tt{Wheel}} data was modulated by a lithography mask in CACTI system \cite{llull2013coded} with $c=14$. Reconstructing these real captured measurements is very challenging due to noise effects. Besides, the masks used in these systems are not ideal binary due to nonuniform illumination. Despite this challenging setting, our method still provides decent reconstruction results with only one training. Fig.~\ref{fig:real} clearly demonstrates that SPA-DUN has sharper edges in {\tt{Domino}}, fewer artifacts in {\tt{Hand}}, and more details without over-smoothing in {\tt{Wheel}}. The above observations show the feasibility and effectiveness of SPA-DUN in real applications.

\section{Conclusion}
In this paper, an efficient Sampling-Priors-Augmented Deep Unfolding Network (SPA-DUN) is proposed for video compressive sensing. This optimization-inspired deep unfolding network has good interpretability and reconstruction performance. Benefiting from the designed lightweight backbone network, SPA-DUN achieves the state-of-the-art reconstruction accuracy with lower model complexity, calculation speed and memory consumption. Furthermore, SPA-DUN has excellent generality and robustness benefiting from the proposed SPA-Learning. This means that one single SPA-DUN can handle arbitrary sampling settings without retraining. This great efficiency and generality promotes the real-world application of VCS systems. In the future, we will further extend our SPA-DUN to other image inverse problems.

\bibliographystyle{ACM-Reference-Format}
\bibliography{ref}

\appendix

\end{document}